\documentclass[10pt,twocolumn,letterpaper]{article}

\usepackage{cvpr}
\usepackage{times}
\usepackage{epsfig}
\usepackage{graphicx}
\usepackage{amsmath}
\usepackage{amssymb}
\usepackage{multirow}

\usepackage[pagebackref=true,breaklinks=true,letterpaper=true,colorlinks,bookmarks=false]{hyperref}

\cvprfinalcopy 

\def\cvprPaperID{2089} 

\ifcvprfinal\fi
\begin{document}

\title{Action Recognition in Video Using Sparse
Coding and Relative Features}

\author{Anal\'i Alfaro\\
\normalsize P. Universidad Catolica de Chile\\
\normalsize Santiago, Chile\\
{\tt\small ajalfaro@uc.cl}
\and
Domingo Mery\\
\normalsize P. Universidad Catolica de Chile\\
\normalsize Santiago, Chile\\
{\tt\small dmery@ing.puc.cl}
\and
Alvaro Soto\\
\normalsize P. Universidad Catolica de Chile\\
\normalsize Santiago, Chile\\
{\tt\small asoto@ing.uc.cl}
}

\maketitle

\begin{abstract}
This work presents an approach to category-based action recognition in video 
using sparse coding techniques. The proposed approach includes two main 
contributions: i) A new method to handle intra-class variations by decomposing 
each video into a reduced set of representative atomic action acts or 
key-sequences, and ii) A new video descriptor, ITRA: Inter-Temporal Relational 
Act Descriptor, that exploits the power of comparative reasoning to capture 
relative similarity relations among key-sequences. In terms of the method 
to obtain key-sequences, we introduce a loss function that, for each video, 
leads to the identification of a sparse set of representative key-frames 
capturing both, relevant particularities arising in the input video, as well as 
relevant generalities arising in the complete class collection. In terms of the 
method to obtain the ITRA descriptor, we introduce a novel scheme to quantify 
relative intra and inter-class similarities among local temporal patterns 
arising in the videos. The resulting ITRA descriptor demonstrates to be highly 
effective to discriminate among action categories. As a result, the proposed 
approach reaches remarkable action recognition performance on several popular 
benchmark datasets, outperforming alternative state-of-the-art techniques by a 
large margin.
\end{abstract}

\vspace{-0.5 cm}
\section{Introduction}
\label{Sec:Introduction}
This work presents a new method for 
action recognition in video that incorporates two novel ideas: (1) A new method 
to 
select relevant key-frames from each video, and (2) A
new method to extract an informative video descriptor. In terms of our technique 
for key-frame selection, previous works have also 
built their action recognition schemes on top of key-frames 
\cite{Zao08EntropyKeyFrames}, Snippets \cite{Schindler08Snippets}, Exemplars 
\cite{Weinland:Boyer:2008}, Actoms \cite{Gaidon11Actom}, or other informative 
subset of short video sub-sequences 
\cite{Niebles10TemporalSegments}\cite{Raptis12ActionParts}. As a relevant 
advantage, by representing a video using a compressed set of distinctive 
sub-sequences, it is possible to eliminate irrelevant or noisy temporal patterns 
and to reduce computation, while still retaining enough information to 
recognize a target action \cite{Assa:EtAl:2005}. Furthermore, it is possible to 
obtain a normalized video representation that avoids distracting sources of 
intra-class variation, such as different velocities in the execution of an 
action. 

Previous works have mainly defined a set of key-frames using manual labelling 
\cite{Gaidon11Actom}, clustering techniques 
\cite{Zao08EntropyKeyFrames}, or discriminative approaches 
\cite{Raptis12ActionParts}. In the case of clustering techniques, the 
usual loss functions produce a set of key-frames that captures temporal action 
patterns occurring frequently in the target classes. As a relevant drawback, 
training instances presenting less common patterns are usually poorly 
represented \cite{Zhu:Ramanan:14} and, as a consequence, the diversity of 
intra-class 
patterns is not fully captured. In the case of discriminative approaches, 
identification of relevant key-frames is usually connected to classification 
stages, focusing learning on mining patterns that capture relevant inter-class 
differences. As a relevant drawback, the mining of key-frames again does not 
focus directly on effectively capturing the diversity of intra-class patterns 
that usually arise in complex action videos.

In contrast to previous work, our method to select key-frames explicitly 
focuses on an effective mining of relevant intra-class variations. As a 
novel guiding strategy, our technique selects, from each training 
video, 
a set of key-frames that balances two main objectives: (i) They are informative 
about the target video,
and (ii) They are informative about the complete set of videos in an action 
class. In other words, we simultaneously favour
the selection of relevant particularities arising in the input video, as well
as meaningful generalities arising in an entire class collection. To achieve 
this, we
establish a loss function that selects from each video a sparse set 
of key-frames that minimizes the
reconstruction error of the input video and the complete set of  
videos in the corresponding action class. 

In terms of our technique to obtain an informative video descriptor, most 
current video descriptors are based on quantifying the absolute 
presence or absence of a set of visual features. Bag-of-Words schemes are a 
good example of this strategy \cite{Jurie:Triggs:05}. As a relevant 
alternative, recent works have 
shown that the relative strength \cite{YagnikSRL11}, or similarity among 
visual features \cite{Parikh:Grauman:13}, can be a powerful cue to perform 
visual 
recognition. As an example, the work in 
\cite{YagnikSRL11} demonstrates a notable increase in object recognition 
performance by using the relative ordering, or rank, among feature 
dimensions. Similarly, the work in 
\cite{Kumar09attributeand} achieves excellent results using a feature 
coding strategy based on similarities among pairs of attributes (similes). 

In contrast to previous work, our method to obtain a video descriptor is based 
on quantifying relative intra and inter-class similarities among local 
temporal patterns or key-sequences. As a building block, we use our proposed 
technique to identify key-frames that are augmented with neighbouring frames to 
form local key-sequences encoding local action acts. These key-sequences, 
in conjunction with sparse 
coding techniques, are then used to learn temporal class-dependent 
dictionaries of local acts. As a key idea, cross-projections of acts into 
dictionaries 
coming from different temporal positions or action classes allow us to quantify 
relative 
local similarities among action categories. As we demonstrate, these 
similarities prove to be highly discriminative to perform action recognition in 
video.  

In summary, our method initially represents an action in a video as a
sparse set of meaningful local acts or key-sequences. Afterwards, we use these 
key-sequences to 
quantify relative local intra and inter-class similarities by projecting the 
key-sequences to a bank of dictionaries encoding patterns from 
different temporal positions or actions classes. These similarities form our
core video descriptor that is then fed to a suitable 
classifier to access action recognition in video. Consequently,
this work makes the following three main contributions:
\vspace{-0.2cm}
\begin{itemize}
 \item A new method to identify a set of relevant key-frames in a video that
manages intra-class variations by preserving essential temporal intra-class
patterns. 
\vspace{-0.2cm}  
 \item A new method to obtain a video descriptor that quantifies relative local 
temporal similarities among local action acts. 
 \vspace{-0.2cm}
\item  Empirical evidence indicating that the combination of the two previous 
contributions provides a substantial increase in action recognition performance 
with respect to alternative state-of-the-art techniques. 

\end{itemize}

\vspace{-0.4 cm}
\section{Related Works}
\label{Sec:RelatedWorks}
There is a large list of works related to category-based action 
recognition
in video, we refer the reader to \cite{Aggarwal2011} for a
suitable review. Here, we focus our review on methods that also 
decompose 
the input video into key-sequences, propose related video descriptors, or use 
sparse coding.

\noindent{\bf Key-sequences:}
Several previous works have tackled the problem of action recognition in video 
by representing each video by a reduced set of meaningful temporal 
parts. 
Weiland and Boyer \cite{Weinland:Boyer:2008} propose an action recognition 
approach based on key-frames that they refer to as Exemplars. Schindler and Van 
Gool \cite{Schindler08Snippets} add motion cues by studying the amount of 
frames, or Snippets, needed to recognize periodic human actions. Gaidon et al. 
\cite{Gaidon11Actom} present an action recognition 
approach that is built on top of atomic action units, or Actoms. As a relevant 
disadvantage, at training time, these previous methods require a manual 
selection or 
labelling of a set of key-frames or key-sequences. 

Discriminative approaches to identify key-frames have also been used. Zhao and 
Elgammal \cite{Zao08EntropyKeyFrames} use an 
entropy-based score to select as key-frames the most discriminative frames from 
each video. Liu et al. \cite{Liu13BoostedKeyFrames} propose a method to select 
key-frames using the Adaboost classifier to identify highly discriminative 
frames for each target class. Extending DPMs 
\cite{Felzenszwalb2010a} to action recognition, Niebles et al. 
\cite{Niebles10TemporalSegments} represent a video using global information 
and short temporal motion segments. Raptis and Sigal 
\cite{Raptis13KeyFramingPoselets} use a video frame representation based on 
max-pooled Poselet \cite{Bourdev2010} activations, in conjunction with a latent 
SVM approach to select relevant key-frames and learn action classifiers. In 
contrast to these previous approaches, we do not assume that all videos in an 
action 
class share a common set of key-sequences. In our case, we 
adaptively identify in each video key-sequences that consider reconstruction 
error and similarities to other local temporal patterns present in the class 
collection.

\noindent{\bf Video descriptors:}
Extensive research has been oriented to propose 
suitable spatio-temporal low-level features 
\cite{Laptev03STIP,Dollar05STfeatures, 
Klasser08HOG3D,Laptev08BagFeatures,Liu08Spin,Yu10STF,WangMBH:IJCV:13}. In our 
case, we build our descriptor 
on top of key-sequences that are
characterized by low-level spatio-temporal features. In this sense, the
proposed descriptor is more closely related to mid-level 
representations, such as the 
ones described in \cite{Liu11Attributes,Corso12ActionBank,Wang2011}. In 
contrast to our approach, current mid-level representations do not 
encode local temporal similarities among key-sequences. 

In terms of encoding similarities among training instances, Kumar et al. 
\cite{Kumar09attributeand} propose a method that exploits facial similarities 
with respect to a specific list of reference people. Yagnik et al. 
\cite{YagnikSRL11} presents a locality sensitive hashing approach that provides 
a feature representation based on relative rank ordering. Similarly, Parikh and 
Grauman \cite{Parikh:Grauman:13} use a max-margin approach to learn a 
function that encodes relative rank ordering. Wang et al. \cite{WangFH:13} 
present a 
method that uses 
information about object-class similarities to train a classifier that responds 
more strongly to examples of similar categories than to examples of dissimilar 
categories. These previous works share with our approach the idea of explicitly 
encoding the 
relative strength of visual properties to achieve 
visual recognition. However, they are not based on sparse 
coding, or they do not exploit relative temporal relations among 
visual patterns. 

\noindent{\bf Sparse Coding:}
A functional approach to action recognition is to create dictionaries based on
low-level representations. Several methods can be used to produce a 
suitable dictionary, BoW \cite{Gaidon11Actom,Laptev08BagFeatures,Liu08Spin}, 
Fisher vectors
\cite{oneataFisher:2013,Bernard:2012}, random forest
\cite{Yao10Hough,Yu10STF}, and sparse coding techniques
\cite{Guha12LearningSparseRep,Guo10TunnelsCovariance,Tran11BodyParts,
Castrodad12MotionImagery}. Tran et al. \cite{Tran11BodyParts} use motion 
information from human body parts 
and sparse coding techniques to classify human actions in video. For each body 
part, they build a dictionary  that integrates information from all classes. 
Similarly to our approach, the atoms in each 
dictionary are given by the training samples themselves. As a main drawback, at 
training time, this method requires manual annotation of human body parts. Guha 
and Ward \cite{Guha12LearningSparseRep} explore several schemes to construct an 
overcomplete dictionary from a set of spatio-temporal descriptors extracted from 
training videos, however, this method does not use key-sequences or relative 
features in its operation. Castrodad et al. \cite{Castrodad12MotionImagery} 
propose a hierarchical two-level sparse coding approach for action recognition. 
In contrast to our approach, this work uses a global representation that 
discards local temporal information. Furthermore, it does not exploit key-frames 
or intra-class relations. 

\vspace{-0.1cm}

\section{Our Method}
\label{Sec:Method}
\vspace{-0.1cm}
Our proposed method has three main
parts: i) Video Decomposition, ii) Video Description, and iii) Video 
Classification. We explain next the main details behind each of these parts.

\vspace{-0.1cm}
\subsection{Video Decomposition} \label{videoDecomposition}
\begin{figure}[t]
\begin{center}
\includegraphics[width=\linewidth]{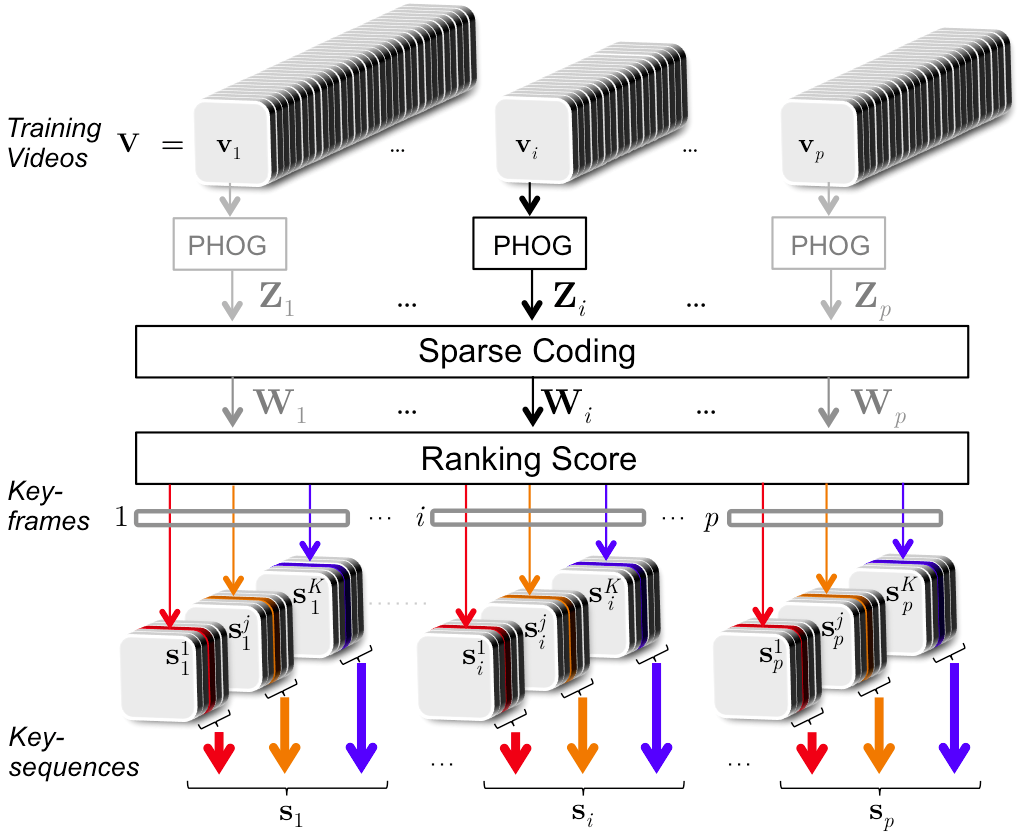}
\caption{Overview of the proposed method to extract key-sequences from an
input video.}
\label{fig:Decomposition}
\end{center}
\end{figure}

\vspace{-0.2cm}
Fig. \ref{fig:Decomposition} summarizes the main steps to decompose an 
input video into a set of $K$ key-sequences. We explain next the details.

\vspace{-0.3cm}
\subsubsection{Selection of Key-Frames}
\vspace{-0.2cm}
We address the selection of key-frames from 
an action video as a reconstruction problem using sparse
coding techniques \cite{Donoho:Elad:03}. Let $ {\bf V} = \{ {\bf v}_i \}_{i=1}^p$ be a
set of $p$ training videos of a given action class, where video ${\bf v}_i$ 
contains $n_i$ frames ${\bf f}_{i}^{j}$, $j \in [1 \dots n_i]$. We encode
each frame ${\bf f}_{i}^{j}$ using  a
pyramid of histograms of oriented
gradients or PHOG--descriptor \cite{Bosch07PHOG}. Then, video ${\bf v}_i$ is 
represented by a matrix ${\bf Z}_i \in \mathbb{R}^{m 
\times
n_i}$, where column $j$ contains the $m$-dimensional PHOG--descriptor 
of frame ${\bf f}_{i}^{j}$.

Our sparse coding representation
considers two main
design goals. First,
similarly to \cite{Elhamifar12KeyFrames}, the atoms of the resulting 
representation
must correspond to frames from the input video. Second, as
mentioned before, the resulting atoms
must simultaneously provide a suitable representation of the input
video and the complete class. To achieve this, for each input video we solve 
the following optimization:
\begin{align}
   \underset{{\bf W}_i,{\bf W}_{(-i)}}{\min}& \| {\bf Z}_i - {\bf Z}_i {\bf
W}_i
\|_F^{2} + \alpha \| {\bf Z}_{(-i)} - {\bf Z}_i {\bf W}_{(-i)} \|_F^{2}
\label{optimization} \\
         \mbox{ s.t. } & \| {\bf W}_i \|_{1,2} \leq \lambda, \nonumber\\ 
         & \| {\bf W}_{(-i)} \|_{1,2} \leq \lambda, \nonumber
\end{align}

\noindent where ${\bf W}_i \in \mathbb{R}^{n_i \times n_i}$ corresponds to the 
matrix 
of coefficients that minimize the constrained reconstruction of the $n_i$ frame 
descriptors in ${\bf Z}_i$. ${\bf Z}_{(-i)} = [\dots, {\bf Z}_{i-1}, {\bf 
Z}_{i+1}, \dots] \in \mathbb{R}^{m \times (n-n_i)}$ corresponds to the matrix of 
PHOG 
descriptors for all the $n$ frames in a target class, 
excluding the $n_i$ frames from video 
${\bf v}_i$. ${\bf W}_{(-i)} = [\dots, {\bf W}_{i-1}, {\bf W}_{i+1}, \dots] \in 
\mathbb{R}^{n_i \times (n-n_i)}$ corresponds to the sparse representation of 
${\bf 
Z}_{(-i)}$ using the frame descriptors in ${\bf Z}_i$. The mixed $\ell_1 /      
\ell_2$ norm is defined as $\| {\bf A} \|_{1,2} \triangleq 
\sum_{i=1}^{N} \| {\bf a}_i \|_2$, where ${\bf A}$ is a sparse matrix and ${\bf 
a}_i$ denotes the $i$-th  row of ${\bf A}$. Then, the mixed norm expresses the 
sum of the $\ell_2$ norms of the rows of ${\bf A}$. Parameter $\lambda  > 0$ 
controls the level of sparsity in the 
reconstruction, and parameter $\alpha > 0$ balances the penalty between errors 
in the reconstruction of video ${\bf v}_i$ and errors in the reconstruction of 
the remaining videos in the class collection. Following 
\cite{Elhamifar12KeyFrames}, we solve the constrained optimization in Eq. 
\ref{optimization} using the {\em
Alternating Direction Method of Multipliers (ADMM)} technique 
\cite{Gabay76ADMM}. 

\begin{figure}
\begin{center}
\includegraphics[width=0.78\linewidth]{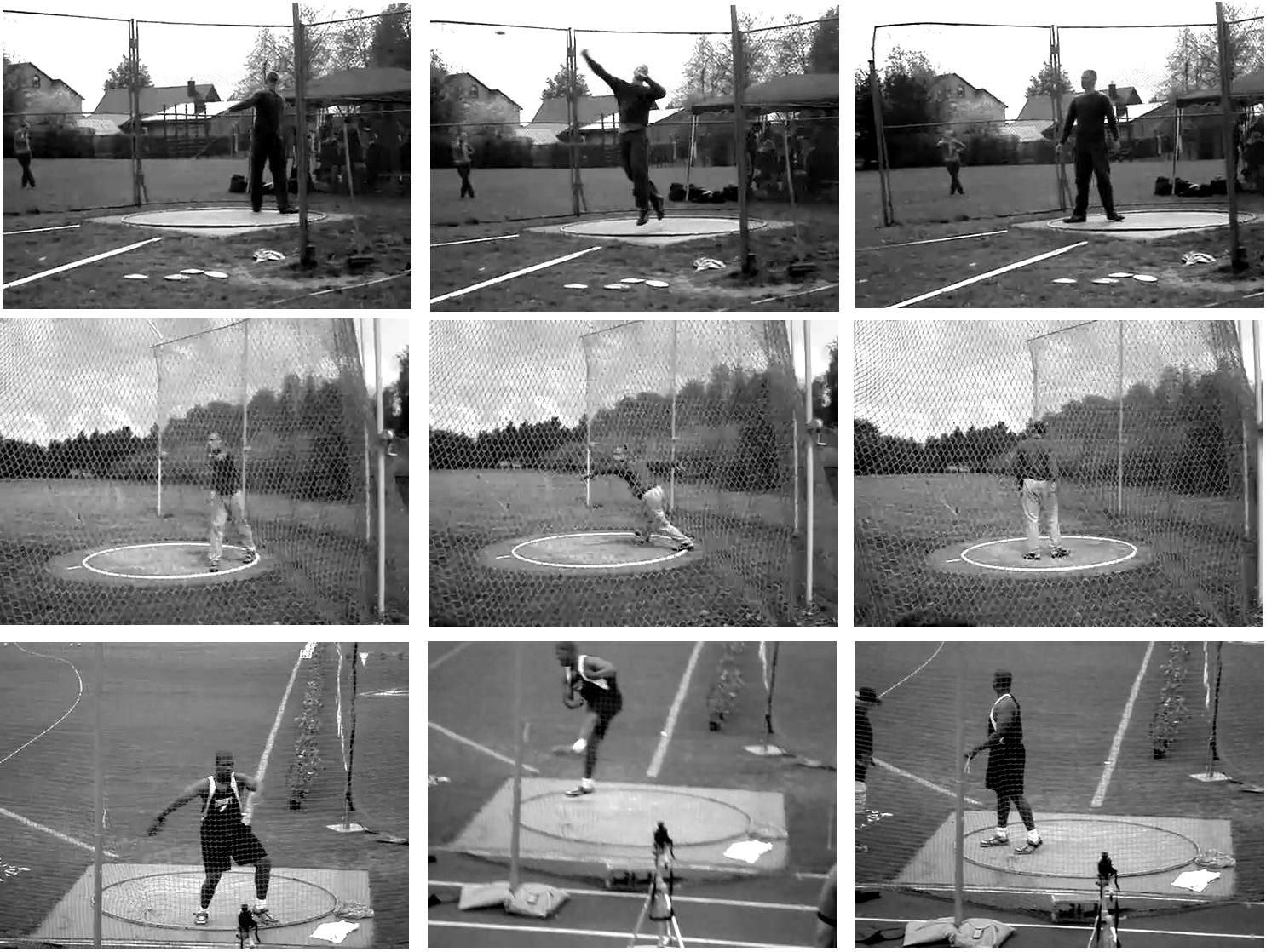}
\end{center}
\vspace{-2mm}
\caption{Key-frames selected by the 
proposed method (rows) for videos of the 
action category \textit{Discus throwing} in the Olympic dataset using K=3.}
\label{Fig:KeySequences}
\vspace{-4mm}
\end{figure}

\vspace{-0.3cm}
\subsubsection{Selection of Key-Sequences}
\begin{figure}
\begin{center}
\includegraphics[width=8cm]{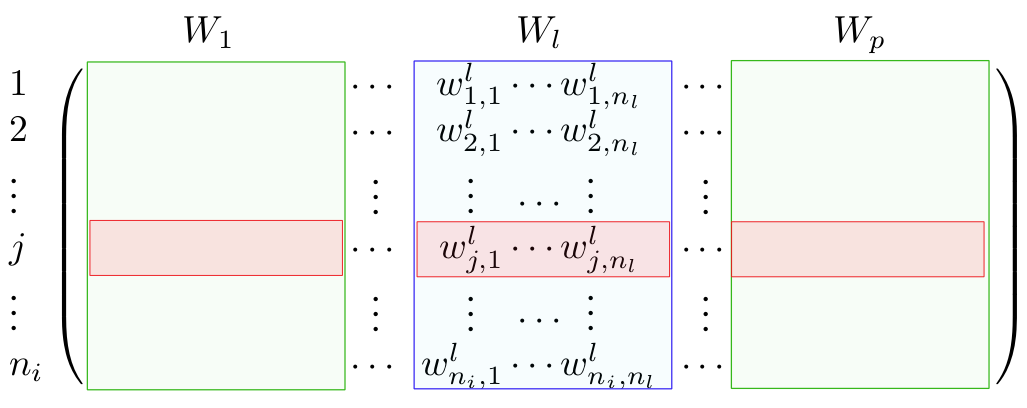}
\end{center}
\vspace{-2mm}
\caption{Matrix ${\bf W^i}$. Columns of ${\bf W^i}$ can be decomposed according 
to the $p$
videos in an action class: ${\bf W^i} = [W_1,\cdots, W_p$]. This
decomposition highlights that each
row $j$ in a submatrix $W_l$ contains information about the contribution
delivered by frame $j$ in video ${\bf v}_i$ to reconstruct the frames in video
${\bf v}_l$. Then, row $j$ in matrix ${\bf W^i}$ contains 
all the reconstruction coefficients associated to frame $j$ in video $v_i$.}
\label{Fig:Wmatriz}
\end{figure}

\vspace{-0.2cm}
Matrix ${\bf W^i} = [ {\bf W}_{i} 
| {\bf W}_{(-i)} ]$ provides
information about the contribution of each
frame in ${\bf v}_i$ to summarize each of the videos in the entire class
collection. Fig. \ref{Fig:Wmatriz} shows a diagram of
matrix ${\bf W^i}$ that highlights this property. Specifically, each row
$j$ in ${\bf
W^i}$ provides information about the contribution provided by frame $j$ in
video ${\bf v}_i$, ${\bf f}_i^j$, to reconstruct the $p$ videos in the class
collection.
Using this property and the notation in Fig. \ref{Fig:Wmatriz}, we define the
following score to quantify the
contribution of frame ${\bf f}_i^j$ to the reconstruction process:
\begin{equation} 
\text{R}({\bf f}_i^j) =  \sum_{l=1}^{p} \sum_{s=1}^{n_l} w_{j,s}^l.  
\end{equation}
\noindent R$({\bf f}_i^j)$ corresponds to the sum of the elements in
the $j$-th row
of matrix $\bf W^i$. We use this score to rank the frames in video ${\bf v}_i$ 
according
to their contribution to the reconstruction process. In particular, a frame 
with a high ranking
score
provides a high contribution
to the reconstruction of the videos in the class collection. Therefore, high
scoring frames represent good candidates to
be selected as key-frames for video ${\bf v}_i$.

Let ${\bf L}_i$ be the set of frames ${\bf 
f}_i^j$ from ${\bf v}_i$ that satisfy R$({\bf 
f}_i^j)> \theta$, where $\theta$ is a given threshold. We
obtain a set of key-frames from ${\bf v}_i$ by selecting $K$ frames from the
candidates in ${\bf L}_i$. Several criterion can be used to select these $K$
frames. In particular, to guarantee that the selected key-frames provide 
a good temporal coverage of the input video, we use the following scheme. First,
we select $K$ time instants uniformly distributed with respect to the length of 
the video. Then, for each of these time instants, we select as a 
key-frame the closest neighbouring frame in ${\bf L}_i$. Fig. 
\ref{Fig:KeySequences} shows instances of 
key-frames selected by this approach using $K=3$.

To include motion cues, we add neighbouring frames to each key-frame in order 
to form 
brief video acts that we refer to as key-sequences. Specifically, for a 
key-frame ${\bf f}_i^j$ in video ${\bf v}_i$, its corresponding key-sequence is 
given by the set ${\bf s}_i^j = \{ {\bf f}_i^l \}_{l=j-t}^{l=j+t}$, i.e., $2t+1$ 
consecutive frames centered at the corresponding key-frame ($t \in  
\mathbb{N}$). Consequently, each 
input video ${\bf v}_i$ is decomposed into a set ${\bf s}_i = \{ {\bf s}_i^1, 
\dots, {\bf s}_i^K \}$, corresponding to $K$ temporally ordered key-sequences.

\subsection{Video Description}
\begin{figure}[t]
\begin{center}
\includegraphics[width=\linewidth]{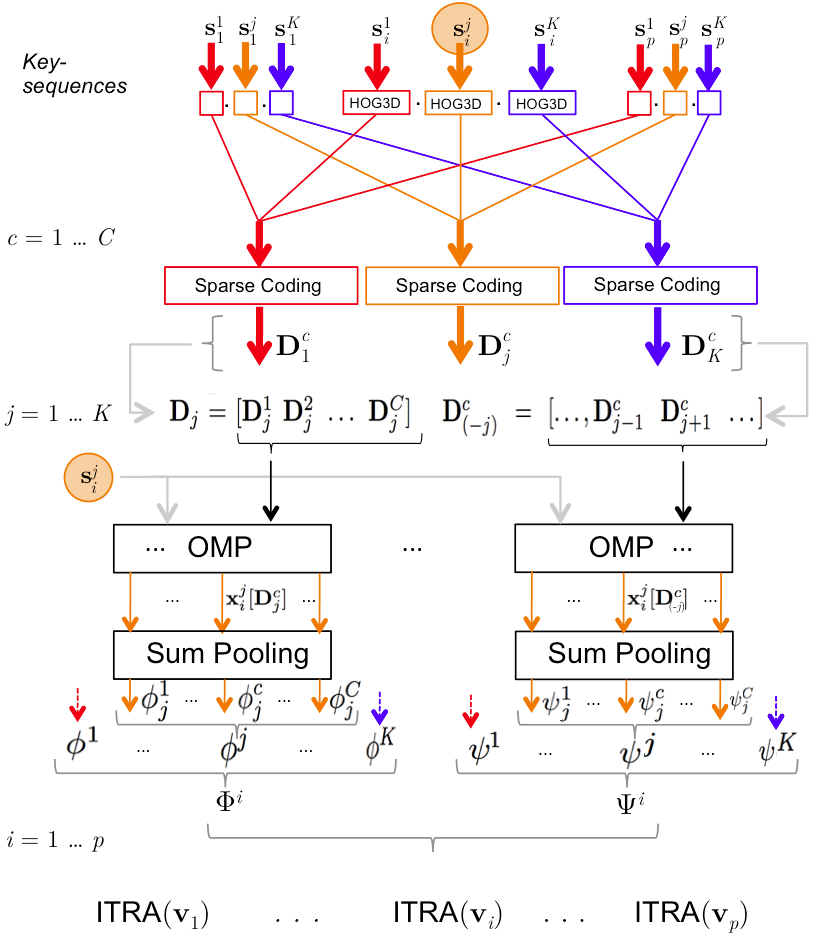}
\end{center}
\vspace{-2mm}
\caption{Overview of the method to obtain the ITRA descriptor. See 
Section \ref{videoDescription} for details.}
\label{fig:ITRA}
\vspace{-2mm}
\end{figure}

\vspace{-0.2cm}
\label{videoDescription}
Fig. \ref{fig:ITRA} summarizes the main steps to build our video descriptor. We 
explain next the details. 

\vspace{-0.3cm}
\subsubsection{Relative Local Temporal Features}
\label{RelativeLTF}
\vspace{-0.2cm}
At the core of our method to obtain relative features is the use
of sparse coding to learn a set of dictionaries that encode
local temporal patterns present in the action classes. Specifically, in the 
case of $C$ action classes and $K$ temporal key-sequences, we
use training data to learn a total of $C \times K$ dictionaries, where
dictionary $D_{k_j}^{c_i}$, $c_i \in [1 \dots C]$, $k_j \in 
[1 \dots K]$, encodes relevant local temporal patterns occurring in class 
$c_i$ at time instance $k_j$.  
 
As a key observation, by projecting a given key-sequence to a concatenated 
version of a subset of the dictionaries, it is possible to quantify the relative 
similarity between the key-sequence and the individual dictionaries. This can be 
achieved by quantifying the total contribution of the atoms in each individual 
dictionary to the reconstruction of the projected key-sequence. As an example, 
consider the case of a concatenated dictionary that encodes local patterns 
learnt from sport actions. In this case, key-sequences from an action class such 
as \textit{running} should use in their reconstruction a significant amount of 
dictionary atoms coming from similar action classes, such as \textit{jogging} 
and \textit{walking}. As a consequence, by quantifying the cross-talk among 
reconstruction contributions coming from different dictionaries, one can obtain 
a feature vector that encodes relative local similarities between the projected 
key-sequence and the temporal patterns encoded in each dictionary. Next, we 
exploit this property to apply two concatenation strategies that allow us to 
obtain a video descriptor capturing inter and   intra-class similarity 
relations.

\vspace{-0.3cm}
\subsubsection{Inter-class Relative Act Descriptor}
\label{InterClassDescriptor}
\vspace{-0.2cm}
Our method to obtain inter-class relative local temporal features is composed of
three
main steps. In the first step we obtain a low-level feature
representation for each key-sequence. Specifically, we randomly sample a set of
spatio-temporal cuboids (300 in our experiments) from each
key-sequence. These cuboids are encoded using the spatio-temporal HOG3D
descriptor
\cite{Klasser08HOG3D}. 
Section \ref{Sec:Experiments} provides further implementation details.

In the second step we use the resulting HOG3D descriptors
and sparse coding to build a set of local temporal dictionaries for each class. Temporal
locality is given by organizing the key-sequences according to their
$K$ temporal positions in the training videos. Let ${\bf Y}^c_j$
be the set of HOG3D descriptors extracted from all key-sequences
occurring at the $j$-th temporal position in the training videos from
class $c$, where $j \in [1,\dots,K]$, $c \in [1,\dots, C]$. We find a 
class-based temporal dictionary ${\bf
D}_j^c$ for position $j$ using the K-SVD algorithm \cite{Aharon06KSVD} to solve:
\begin{equation}
	  \min_{{\bf D}^c_j,{\bf X}^c_j} \| {\bf Y}^c_j - {\bf D}^c_j {\bf
X}^c_j \|_{\sf F}^{2}  \;\; \mbox{ s.t. } \;\; \|  {\bf x}_i \|_{0} \leq
\lambda_1,
\label{funcional3}
\end{equation}
\noindent where ${\bf Y}^c_j \in \mathbb{R}^{m \times n_s}$, $m$ is the 
dimensionality
of the descriptors and $n_s$ is the total number of cuboids sampled from videos
of class $c$ and temporal position $j$, ${\bf D}^c_j \in \mathbb{R}^{m \times
n_a}$, ${\bf X}^c_j \in \mathbb{R}^{n_a
\times n_s}$, $n_a$ is the number of atoms in each dictionary ${\bf D}^c_j$, and
the sparsity restriction on each column ${\bf x}_i \in {\bf X}^c_j$ indicates
that its total number of nonzero entries must not exceed $\lambda_1$.

Finally, in the third step we use the previous set of dictionaries
to obtain a local temporal similarity descriptor for each key-sequence. To achieve this,
for each temporal position $j$, we concatenate the $C$ class-based dictionaries obtained 
in the previous step. This provides a set of $K$ temporal dictionaries, where each dictionary 
contains information about local patterns occurring in all target classes at a
given
temporal position $j$. These $K$ representations allow us to quantify local temporal 
similarities among the target classes. Specifically, let ${\bf D}_j = [{\bf
D}^1_j \ {\bf
D}^2_j \ \dots \ {\bf D}^C_j ]$ be the concatenated temporal dictionary
corresponding to temporal position $j$. To obtain a descriptor for
key-sequence ${\bf s}_i^j$ from video ${{\bf v}_i}$, we first project ${\bf
s}_i^j$ onto dictionary ${\bf
D}_j$ imposing a sparsity constraint. We achieve this by using the Orthogonal
Matching Pursuit (OMP) technique to solve:
\begin{equation}
    \min_{{\bf x}_i^j} \| {\bf s}_i^j - {\bf D}_{j} {\bf x}_i^j
\|_{\sf F}^{2}  \;\; \mbox{ s.t. } \;\; \| {\bf x}_i^j\|_{0} \leq \lambda_2,
       \label{funcional4}
\end{equation}
\noindent where vector ${\bf x}_i^j=\{{\bf x}_i^j[{\bf D}_j^1], \dots, {\bf
x}_i^j[{\bf D}_j^C]\}$ is the resulting set of coefficients, and a component
vector ${\bf x}_i^j[{\bf D}_j^c] \in \mathbb{R}^{n_a}$ corresponds to the 
coefficients
associated to the projection of ${\bf s}_i^j$ onto the atoms in
subdictionary ${\bf D}_j^c$.

We quantify the similarity of ${\bf s}_i^j$ to the atoms
corresponding to each class by using a sum-pooling operator that 
evaluates the contribution provided by the words in each subdictionary ${\bf
D}_j^{c}$ to the reconstruction of ${\bf s}_i^j$. We
define this sum-pooling operator as:
\begin{equation} \label{sumPooling}
  \phi^c_j ({\bf s}_i^j)= \sum_{l=1}^{n_a} {\bf x}_i^j[{\bf
D}_j^c](l).
\end{equation}

By applying the previous method to the set of $K$ key-sequences ${\bf s}_i^j$ in
a video ${{\bf v}_i}$, we obtain
a video descriptor ${\bf
\Phi}^i = [ \phi^1 , \dots, \ \phi^K ] \in
\mathbb{R}^{C \times K}$, where each component vector $\phi^j$ is given by 
$\phi^j=[\phi^1_j, \dots,
\phi^C_j]$. In this way, ${\bf
\Phi}^i$ contains information about relative {\em inter-class} similarities
among
key-sequences or acts. Therefore, we refer to this descriptor as {\em
Inter-class Relative Act Descriptor}. 

\vspace{-0.3cm}
\subsubsection{Intra-class Relative Act Descriptor}
\label{IntraClassDescriptor}
\vspace{-0.2cm}
The procedure in Section \ref{InterClassDescriptor} provides a descriptor 
that encodes relative local
temporal similarities across the target classes. In this section, we use a
similar
procedure to obtain local temporal similarities at an intra-class level. 
Specifically, we quantify the similarity of a key-sequence occurring at
temporal position $j$ with respect to the patterns occurring at the remaining $K-1$ temporal
positions in a target class. To do this, we follow the procedure described in 
Section \ref{InterClassDescriptor} but,
this time we project a key-sequence ${\bf s}_i^j$ onto the concatenated
dictionary ${\bf
D}_{(-j)}^{c} = [\dots, {\bf D}^{c}_{j-1} \; {\bf D}^{c}_{j+1} \ \dots]$,
i.e., the concatenation of the $k-1$ key-sequence dictionaries for class $c$,
excepting the dictionary corresponding to temporal position
$j$. We again use the OMP technique to perform this projection, i.e., to
solve:
\begin{equation}
    \min_{{\bf x}_i^j} \| {\bf s}_i^j - {\bf D}_{(-j)}^{c} {\bf x}_i^j
\|_{\sf F}^{2}  \;\; \mbox{ s.t. } \;\; \| {\bf x}_i^j\|_{0} \leq \lambda_3.
       \label{funcional5}
\end{equation}
Similarly to 
the Inter-class Relative Act Descriptor, 
we obtain a video
descriptor, ${\bf\Psi}^i = [ \psi^1 , \dots, \ \psi^K ] \in \mathbb{R}^{K \times
(K-1)}$, 
by applying the projection to all key-sequences in a video ${\bf v}_i$ and then 
using the
corresponding sum-pooling operations to quantify the reconstruction contribution
of each subdictionary ${\bf D}^{c}_{l}$. In this way, ${\bf \Psi}^i$ contains 
information about relative {\em
intra-class}
similarities among key-sequences or local acts, therefore, we refer to this
descriptor as {\em Intra-class Relative Act Descriptor}.

\vspace{-0.3cm}
\subsubsection{Inter Temporal Relational Act Descriptor: {\normalfont
ITRA}}
\label{ItraClassDescriptor}
\vspace{-0.2cm}
We obtain a final feature vector descriptor for a video ${\bf v}_i$ by
concatenating the Inter
and Intra-class Relative Act Descriptors. We refer to this new
descriptor as {\em Inter
Temporal Relational
Act Descriptor} or ITRA, where {\em ITRA}(${{\bf v}_i}$) = $\{ {\bf\Phi}^i
\cup {\bf \Psi}^i \} \in \mathbb{R}^{K
\times (C+(K-1))}$.

\subsection{Video Classification}
\label{SubSec:Classification}
ITRA can be used by any off-the-shelf supervised
classification scheme. Here, we use a sparse coding approach. 
\vspace{-1.0cm}
\paragraph{Training:} During the training phase, we first use the method 
described in Section
\ref{videoDecomposition} to decompose each training video into a set of
key-sequences. Then, we use the method described in Section
\ref{videoDescription} to obtain the ITRA descriptor for each training video.
Afterwards, these descriptors, along with sparse coding techniques, are
used to build a dictionary for each target class. Specifically, let ${\bf Y}^c$ 
be a
matrix containing in its columns the ITRA descriptors corresponding to
the training videos from action class $c \in [1,\dots,C]$. For each action
class, we use the K-SVD algorithm \cite{Aharon06KSVD} to obtain a class-based
dictionary ${\bf B}^c$ by solving:
\begin{equation}
\label{eq_classification1}
\min_{{\bf B}^c,{\bf X}^c} \|{\bf Y}^c - {\bf B}^c {\bf X}^c \|_{\sf F}^{2}
\;\;\; \mbox{
s.t. } \;\;\; \|{\bf x}_i\|_0 \leq  \lambda_4, \;\; \forall i,
\end{equation}
\noindent where ${\bf B} ^c \in \mathbb{R}^{|\mbox{\tiny{ITRA}}| \times n_a}$,
$|\mbox{ITRA}|$ represents the dimensionality of the ITRA descriptor, and $n_a$
is the selected number of atoms to build the dictionary. ${\bf X}^c$
corresponds to the matrix of coefficients and vectors ${\bf x}_i$ to its
columns. As a final step, we concatenate the $C$ class-based dictionaries to 
obtain the
joint dictionary ${\bf B} = [ {\bf B}^1 | {\bf B}^2 | \cdots | {\bf B}^C ]$
that forms the core
representation to classify new action videos.
\vspace{-0.3cm}
\paragraph{Inference:} To classify a new input video, similarly to the training 
phase, we first use the
methods in Sections \ref{videoDecomposition} and \ref{videoDescription} to
obtain its ITRA descriptor. As a relevant difference from the training
phase, in
this case we do not know the class label of the input video, therefore, we need
to obtain its key-sequence decomposition with respect to each target class.
This task
leads to $C$ ITRA descriptors to represent each input video. Consequently, the
classification of an input video consists of projecting these $C$ descriptors
onto the joint dictionary ${\bf B}$ and then using a majority vote scheme to
assign the
video to the class that contributes the most to the reconstruction of the
descriptors. Specifically, let ${\bf v}_q$  be a test video, and $\Omega^c({\bf
v}_q)$ its ITRA descriptor with respect to class $c$. We obtain a 
sparse representation for each of the $C$ ITRA descriptors using the OMP
technique to solve:

\begin{equation}
\label{eq_classification2}
\min_{{\bf \alpha}_c} \|\Omega^c({\bf v}_q) - {\bf B} {\bf \alpha}_c \|_{\sf
2}^{2} \;\; \mbox{ s.t. } \;\;
\|{\bf \alpha}_c\|_0 \leq  \lambda_5.
\end{equation}

The previous process provides $C$ sets of coefficients $\alpha_c$. We use each
of these sets to obtain a partial classification of the input video. We
achieve this by applying, to each set, a sum-pooling
operator similar to the one presented in Eq. (\ref{sumPooling}), and classifying
the input video according to the class that presents the greatest contribution
to the reconstruction of the corresponding set ${\bf
\alpha}_c$. Finally, using
these $C$ partial classifications, we use majority vote to assign the input
video to the most voted class.

\vspace{-0.1cm}
\section{Experiments and Results}
\label{Sec:Experiments}
\vspace{-0.2cm}
We validate our method by using three popular benchmark datasets for 
action recognition: KTH
\cite{Schuldt04recognizinghuman}, Olympic \cite{Niebles10TemporalSegments}, and 
HOHA \cite{Laptev08BagFeatures}. In all the experiments, we select values for 
the main
parameters using the following criteria.

\noindent{\bf Estimation of Key-Sequences:} We use training data
from the Olympic dataset to tune the number of acts to represent an action.
Experimentally, we find that 3 acts are enough to achieve high
recognition performance. Hence, in all our experiments, we select
$K=3$ key-sequences to represent each training or test video. 

In terms of the time span of each key-sequence, we take a fixed group of
7 frames ($t=3$) to form each key-sequence. For each sequence we randomly
extract
$300$ cuboids, described using HOG3D (300 dimensions). To filter out
uninformative cuboids, we set a threshold to the magnitude of the
HOG3D descriptor. We calibrate this threshold to eliminate the
5\% least informative cuboids from each dataset. Afterwards, the remaining
descriptors are normalized. Table \ref{TableThresholds}
shows the value of
the
resulting thresholds for each dataset. 

\begin{table}[h]
\begin{center}
\begin{tabular}{|l|c|c|}
\hline 
 Dataset        & Train  & Test \\
\hline\hline
   KTH          &  2.5  & 2.5  \\ 
   Olympic      &  2  & 2  \\
   HOHA         &  1.3  & 1.6   \\
\hline 

\end{tabular}
\end{center}
\caption{Thresholds used to filter out uninformative cuboids from the
key-sequences. For each dataset, we calibrate this threshold to eliminate the
5\% least informative cuboids.}
\label{TableThresholds}
\end{table}

\noindent{\bf Estimation of ITRA descriptor:} Parameters for the
extraction of ITRA descriptors are
related to the construction of the dictionaries described in Section
\ref{videoDescription}. Let $\mu$ be the redundancy
\footnote{Redundancy indicates the folds of basis vectors that need to be
identified with respect to the dimensionality of the descriptor.} and let
$\delta$ be
the dimensionality of a descriptor. Following the empirical results
in \cite{Guha12LearningSparseRep}, we fix the number of atoms in each local
dictionary to be $n_a = \mu \times \delta$. Therefore, the number of atoms for
the
concatenated dictionaries are: $P = \mu \times \delta \times C$ for the
extraction of \textit{Inter-class Relative Act Descriptors}, ${\bf \Phi}$, and
$P = \mu \times \delta \times (K-1)$ for the extraction of \textit{Intra-class
Relative Act Descriptors}, ${\bf \Psi}$. In our experiments, we use $\mu=2$
and
$\delta=300$. As a result, the dimension of
the ITRA descriptors for KTH, Olympic, and HOHA datasets are $24$, $54$
and $30$, respectively. Also, following \cite{Guha12LearningSparseRep}, the
sparsity parameters $\lambda_1$, $\lambda_2$, and $\lambda_3$, are set to be
10\% of the number of atoms.\\
\noindent{\bf Classifier:} Parameters  $P,
\mu$, $\lambda_4$, and $\lambda_5$ are configured
using the same scheme described above.

\subsection{Action Recognition Performance}
\label{Sec:Comparison}
\noindent{\bf KTH Dataset:}
This set contains 2391 video sequences displaying six types of human
actions. In
our experiments we use the original setup \cite{Schuldt04recognizinghuman} to
divide the data into training and test sets. Table \ref{KTHComparison} shows
the recognition performance reached by our method. Table
\ref{KTHComparison} also includes the performance of alternative action
recognition schemes proposed in the literature, including
approaches that also use sparse coding techniques \cite{Alfaro13Acts,
Castrodad12MotionImagery}. Our method obtains 
a recognition performance of 97.5\%.

\begin{table}[h]
\begin{center}
\begin{tabular}{|l|c|c|}
\hline 
Method & Acc.\\
\hline\hline
Laptev   et al. \cite{Laptev08BagFeatures} (2008) & 91.8\%  \\
Niebles  et al. \cite{Niebles10TemporalSegments} (2010) & 91.3\%  \\
Castrodad et al. \cite{Castrodad12MotionImagery} (2012) & 96.3\%  \\
Alfaro   et al. \cite{Alfaro13Acts} (2013) & 95.7\%  \\
\hline \hline
Our method                                        &    \textbf{97.5\%} \\ 
\hline
\end{tabular}
\end{center}
\caption{Recognition rates of our and alternative methods on KTH 
dataset. In all cases, the same testing protocol is used.}
\label{KTHComparison}
\end{table}

\begin{figure*}[t]
\begin{center}
\includegraphics[width=13cm]{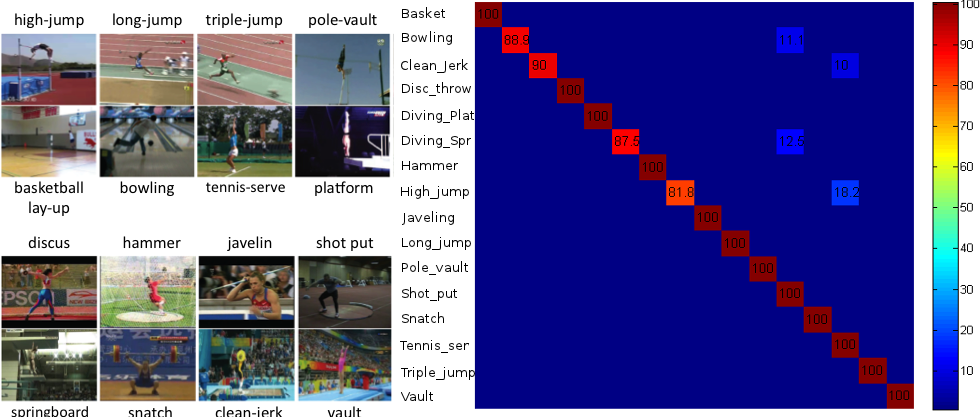}\\
\end{center}
\caption{Olympic dataset. Left: sample from each action class. Right: confusion
matrix for our method on Olympic dataset.}
\label{FigConfusionOlympic}
\end{figure*}

\noindent{\bf Olympic Dataset:} 
This dataset contains 16 actions corresponding to 783 videos of athletes
practicing different sports \cite{Niebles10TemporalSegments}. Fig.
\ref{FigConfusionOlympic} shows sample frames displaying the
action classes. In
our experiments, we use the original setup \cite{Niebles10TemporalSegments} to
divide the data into training and test sets. Table \ref{OlympicComparison}
shows the recognition performance reached by our method and several
alternative
state-of-the-art techniques. Our approach achieves a recognition rate of
96.3\%. This is a remarkable increase in performance with respect to
previous state-of-the-art approaches. Fig. \ref{FigConfusionOlympic} shows the
confusion matrix
reported by our
method. We note that many actions from this dataset have a perfect recognition
rate. Therefore, our approach effectively captures relevant acts
and their temporal relationships. 

\begin{table}[h]
\begin{center}
\begin{tabular}{|l|c|c|}
\hline 
 Method & Acc. \\
\hline\hline
Niebles     et al. \cite{Niebles10TemporalSegments}     (2010) & 72.1\% \\
Liu         et al. \cite{Liu11Attributes}               (2011) & 74.4\% \\
Jiang       et al. \cite{Jiang12MotionReferencePoints}  (2012) & 80.6\% \\
Alfaro      et al. \cite{Alfaro13Acts}		(2013) & 81.3\% \\
Gaidon      et al. \cite{Gaidon:2014}           (2014) & 85.0 \% \\
\hline \hline
Our method                                             & \textbf{96.3\%} \\
\hline
\end{tabular}
\end{center}
\caption{Recognition rates of our and alternative methods on Olympic 
dataset. In all cases, the same testing protocol is used.}
\label{OlympicComparison}
\end{table}

\begin{figure*}[t]
\begin{center}
\includegraphics[width=16cm]{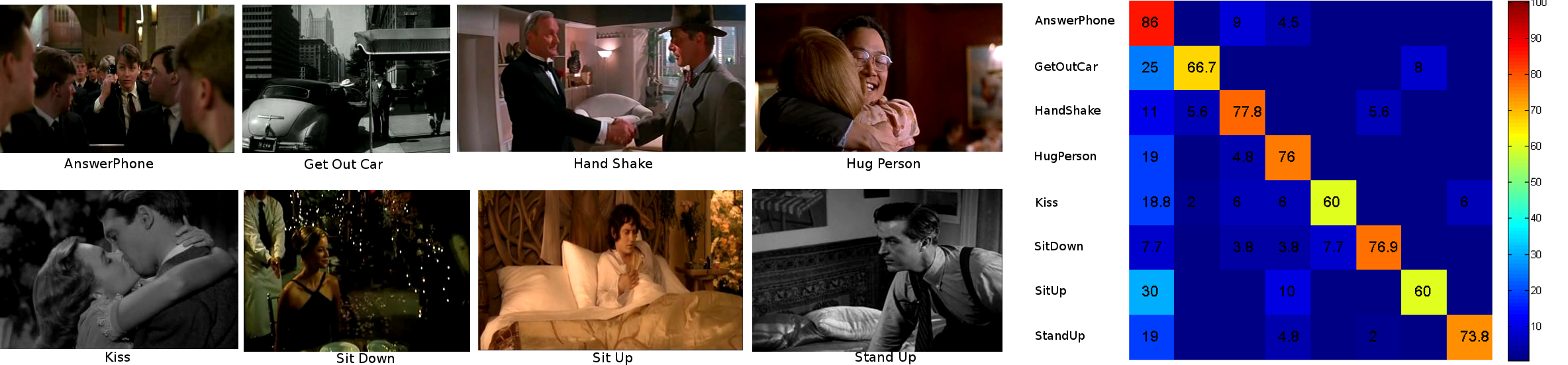}\\
\end{center}
\caption{HOHA. Left: sample from each action class. Right: confusion
matrix for our method on HOHA dataset.}
\label{FigConfusionHOHA}
\end{figure*}

\noindent{\bf Hollywood Dataset:} 
This dataset contains video clips extracted from 32 movies and displaying 8
action classes. Fig.
\ref{FigConfusionHOHA} shows sample frames displaying the
action classes. We use only the videos
with manual annotations (clean training file) and we limit the dataset to
videos with a single label. This is the same testing protocol used by the
alternative techniques considered here. Table \ref{HOHAComparison}
shows the recognition performance of our
method and several alternative
state-of-the-art techniques. Our approach achieves a recognition rate
of 71.9\%. Again, this is a remarkable increase in performance with respect to
previous state-of-the-art approaches. Fig. \ref{FigConfusionHOHA} shows the
confusion matrix
reported by our method. Actions such as
\textit{answer phone}, \textit{handshake}, and \textit{hug person} obtain high
recognition rates. In
contrast, the actions \textit{get out car}, \textit{kiss}, and \textit{sit up}
present a lower recognition performance. According to the confusion matrix in 
Fig. \ref{FigConfusionHOHA}, these
actions present a high confusion rate with respect to the
action \textit{answer phone}. This can be explained by the presence of a common 
pattern among these actions in this dataset, which is given by 
a slow
incorporation of the main actor.

\begin{table}[h]
\begin{center}
\begin{tabular}{|l|c|c|}
\hline 
 Method & Acc.\\
\hline\hline
Laptev   et al. \cite{Laptev08BagFeatures}        (2008) & 38.4\%  \\
Wang     et al. \cite{Wang09STFeatures}          (2009) & 47.4\%  \\
Wu       et al. \cite{Wu11LagrangianParticles}   (2011) & 47.6\%   \\
Zhou     et al. \cite{Zhou:2015}                 (2015) & 50.5\% \\
\hline \hline
Our method                                      & \textbf{71.9\% }\\
\hline
\end{tabular}
\end{center}
\caption{Recognition rates of our and alternative methods on HOHA dataset. In 
all cases, the same testing protocol is used.}
\label{HOHAComparison}
\end{table}

\subsection{Evaluation of Method to Extract Key-Sequences}
\label{Sec:EvaluationKeyFrames}
In this
section, we evaluate the relevance of the proposed method to obtain
key-frames by replacing this step of our approach by
alternative strategies. Besides this modification, we maintain the
remaining steps of our approach and use the same parameter values reported in
Section
\ref{Sec:Comparison}. In particular, we implement two baselines, Table 
\ref{frameComparison} shows 
our results:

\noindent \textbf{Baseline 1 (B1)}, \textit{Uniform Selection}: We split the 
video into 
$K$
equal-sized temporal segments and select the central frame of
each segment as a key-frame. 

\noindent \textbf{Baseline 2 (B2)}, \textit{K-Means}: We generate all possible 
video
sequences
containing $2t+1$ frames by applying a temporal sliding window. We then apply
the K-Means clustering algorithm to each class to obtain $K$
cluster centers per class. For each video, we select as key-frames the most
similar descriptor to each cluster center.

\begin{table}[h]
\begin{center}
\begin{tabular}{|l|c|c|r|}
\hline 
 \multirow{2}{*}{Dataset} & \multicolumn{3}{|c|}{Method} \\
\cline{2-4}
  { }        &   B1    & B2    &  Ours \\
\hline
\hline
  {HOHA}     & 34.2\%  & 37.2\%  & \bf{71.9\%}  \\
\hline
  {Olympic}  & 46.3\%  &  63.4\%  & \bf {96.3\%}  \\
\hline
\end{tabular}
\end{center}
\caption{Performances of our method and alternative strategies
to extract key-sequences.}
\label{frameComparison}
\end{table}
\vspace{-0.2cm}
\subsection{Evaluation of ITRA descriptor}
\label{Sec:Classification}
We evaluate the effectiveness of our ITRA descriptor by replacing this
part of our approach by alternative schemes to obtain a video descriptor.
These alternative schemes are also based on sparse coding techniques but they
do not exploit relative local or temporal
information. Specifically, we consider two baselines, Table 
\ref{ITRAevaluation} shows our results:

\noindent \textbf{Baseline 1 (B1)}, \textit{Ignoring relative local temporal
information}: All key-sequences from all temporal positions are combined to
build a single class-shared joint dictionary that do not preserve temporal
order among the key-sequences. This baseline can
be considered as a BoW type of representation that does not encode relative
temporal relations among key-sequences. 

\noindent \textbf{Baseline 2 (B2)}, \textit{Ignoring intra-class relations}: 
this baseline
only considers the term in ITRA descriptor associated to
the {\em Inter-Class Relative Act Descriptor} ${\bf
\Phi}^i$, discarding intra-class relations provided by the {\em Intra-Class
Relative
Act Descriptor} ${\bf\Psi}^i$.
\begin{table}[h]
\begin{center}
\begin{tabular}{|l|c|c|r|}
\hline 
  \multirow{2}{*}{Dataset} & \multicolumn{3}{|c|}{Method} \\
\cline{2-4}
  { }        &   B1 &   B2    & Ours \\
\hline
\hline
  {HOHA}     &  42.2\%  & 51.3\%   &  \bf{71.9\%}  \\
\hline
  {Olympic}  &  72.4\%  &  87.3\%  &  \bf {96.3\%}  \\
\hline
\end{tabular}
\end{center}
\caption{Performances of our method and alternative strategies to construct the
video descriptor using sparse coding techniques.}
\label{ITRAevaluation}
\end{table}
\vspace{-0.6cm}
\section{Conclusions}
\label{Sec:Conclusion}
\vspace{-0.2cm}
We present a novel method for category-based action recognition in video. As a 
main result, our experiments show that the proposed method reaches remarkable 
action recognition performance on 3 popular benchmark datasets. 
Furthermore, the reduced dimensionality of the ITRA descriptor provides a fast 
classification scheme. Actually, using a reduced 
dimensionality, between 24 and 54 dimensions for the datasets considered here, 
it provides a representation that demonstrates to be highly discriminative. 
As 
future work, the ITRA descriptor opens the possibility to
explore several strategies to concatenate the basic dictionaries
to access different relative similarity relationships. 


\vspace{-0.3cm}
{\small
\section*{Acknowledgements}
\label{sec:acknowledgements}
\vspace{-0.3cm}
This work was partially funded by FONDECYT grant 1151018, CONICYT, Chile.
}

{\small
\bibliographystyle{ieee}
\bibliography{BibliografiaAnali}
}

\end{document}